\documentclass[letterpaper, 10 pt, conference]{ieeeconf}  

\IEEEoverridecommandlockouts                              

\usepackage{graphicx,subfig}

\usepackage{hyperref}
\usepackage{biblatex}
\usepackage{booktabs}  
\usepackage{threeparttable} 
\usepackage{makecell}
\usepackage{graphicx}
\usepackage{multicol}
\usepackage{multirow}
\usepackage{feyn}
\usepackage{amsmath,amssymb,amsfonts}
\usepackage{ulem}
\usepackage{caption}
\usepackage{fancyhdr} 

\usepackage{hyperref}
\hypersetup{
	colorlinks=true,
	linkcolor=cyan,
	filecolor=blue,      
	urlcolor=red,
	citecolor=green,
}

\title{\LARGE \bf
CLRKDNet: Speeding up Lane Detection with Knowledge Distillation
}

\author{Weiqing Qi, Guoyang Zhao, Fulong Ma, Linwei Zheng and Ming Liu
\thanks{This work was supported by Guangzhou-HKUST(GZ) Joint Funding Program (No. 2024A03J0618), awarded to Prof. Ming Liu.
\textit{(Corresponding author: Ming Liu (email:eelium@hkust-gz.edu.cn).)}}
\thanks{Weiqing Qi, Guoyang Zhao, Fulong Ma, Linwei Zheng and Ming Liu are with the Hong Kong University of Science and Technology (Guangzhou), Nansha, Guangzhou, 511400, Guangdong, China.
 \textit{(email:{\{wqiad, gzhao492, fmaaf, lzhengad\}@connect.hkust-gz.edu.cn)}.}}
}

\begin{document}

\maketitle

\thispagestyle{empty}
\pagestyle{empty}

\begin{abstract}


Road lanes are integral components of the visual perception systems in intelligent vehicles, playing a pivotal role in safe navigation. In lane detection tasks, balancing accuracy with real-time performance is essential, yet existing methods often sacrifice one for the other. To address this trade-off, we introduce CLRKDNet, a streamlined model that balances detection accuracy with real-time performance. The state-of-the-art model CLRNet has demonstrated exceptional performance across various datasets, yet its computational overhead is substantial due to its Feature Pyramid Network (FPN) and muti-layer detection head architecture.  Our method simplifies both the FPN structure and detection heads, redesigning them to incorporate a novel teacher-student distillation process alongside a newly introduced series of distillation losses. This combination reduces inference time by up to 60\% while maintaining detection accuracy comparable to CLRNet. This strategic balance of accuracy and speed makes CLRKDNet a viable solution for real-time lane detection tasks in autonomous driving applications. 
\textit{Code and models are available at: \url{https://github.com/weiqingq/CLRKDNet}.}

\end{abstract}


\section{Introduction}

Lane detection \cite{zhang2021deep,ma2024monocular} is a critical aspect of intelligent transportation, including autonomous driving and Advanced Driver Assistance Systems (ADAS). Lanes are integral elements of road traffic, delineating vehicular paths and promoting safer and smoother driving conditions. Detection methods generally fall into two categories: model-based and feature-based. Model-based approaches utilize predefined lane models, interpreting lane recognition as a parameter estimation problem, which helps reduce sensitivity to noise and limits reliance on extensive local image areas. In contrast, feature-based approaches classify individual image points as lane or non-lane, relying on specific characteristics such as edge gradient, width, intensity, and color. This strategy, however, requires distinct lane edges and strong color contrasts for accurate detection. Both approaches follow a similar sequence of steps: region of interest (ROI) extraction, image preprocessing, feature extraction, and lane fitting.

With the advent of deep learning \cite{krizhevsky2012imagenet,ma2023dataset}, traditional lane detection methods have largely been supplanted by more advanced, end-to-end deep learning approaches. These modern methods eliminate the need for manual feature engineering and enhance both the robustness and efficacy of detection systems. Contemporary lane detection techniques can be broadly classified into four categories: segmentation-based, parametric curve-based, keypoint-based, and anchor-based methods.  Segmentation-based approaches \cite{pan2018spatial,zheng2021resa,wang2018lanenet} treat lane detection as a semantic segmentation task, while parametric curve-based methods \cite{feng2022rethinking,tabelini2021polylanenet,liu2021end} model lanes using curve parameters, detecting them through parameter regression.  Keypoint-based methods \cite{qu2021focus} approach lane detection as a problem of keypoint estimation followed by integration. Finally, anchor-based methods \cite{tabelini2021keep,zheng2022clrnet,qin2020ultra,qin2022ultra,liu2021condlanenet,wei2022row} employ line-shaped anchors, regressing offsets of sampled points from these predefined anchors.

\begin{figure}[t]
    \centering
    \captionsetup{font={footnotesize}}
    \includegraphics[width=1.0\linewidth]{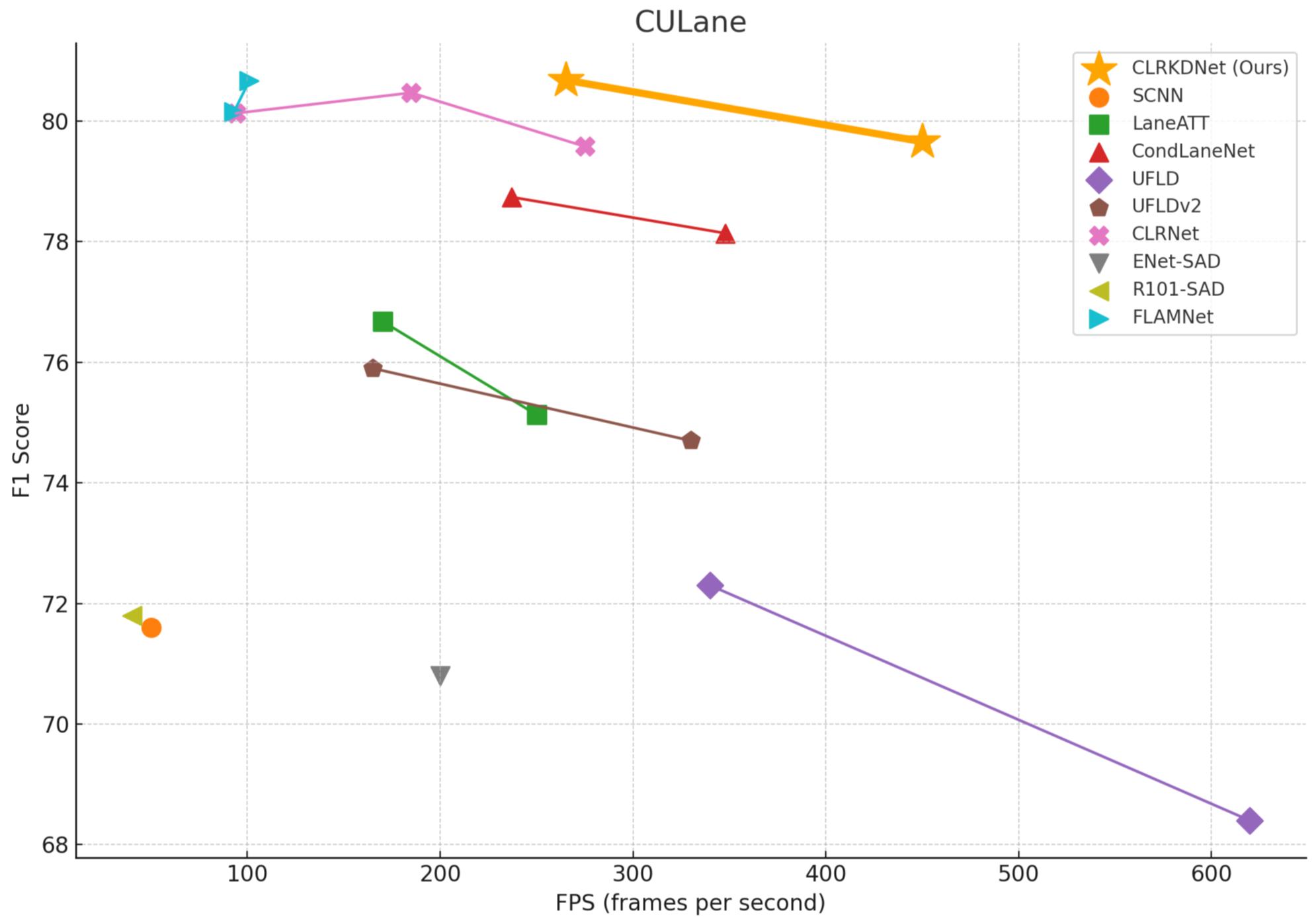}
    \captionsetup{font={footnotesize}}
    \caption{FPS vs. F1-score of state-of-the-art methods on CULane benchmarks.}
    \label{fps_vs_f1}
    \vspace{-15pt}
\end{figure}

Despite considerable advances in lane detection using deep learning techniques, there are still opportunities for further enhancements. Lane lines typically extend over long pixel stretches in images and exhibit distinguishable characteristics from the road surface at local scales, underscoring the need to extract both global and local features for accurate detection. In the paper by Zheng \textit{et al.} \cite{zheng2022clrnet}, the Cross-Layer Refinement Network (CLRNet) was introduced, leveraging both high-level semantic and low-level detailed features. It begins with rough localization using high-level features, followed by refinement with detailed features for precise lane positioning. The ROIGather module further captures extensive global contextual information by linking ROI lane features to the entire feature map, significantly elevating detection performance compared to previous methodologies. However, CLRNet's complexity, including its Feature Pyramid Network (FPN) and multiple detection heads, increases inference time, hindering real-time performance crucial for autonomous vehicles. In response, we developed CLRKDNet, a model designed to reduce inference time while maintaining accuracy. CLRKDNet simplifies the FPN architecture, opting for a streamlined feature aggregation network, and reduces the number of detection heads, thereby eliminating the iterative refinement process. This model uses CLRNet as a teacher model, incorporating a novel knowledge distillation procedure to enhance the performance of the streamlined student model and offset potential declines in detection accuracy. This multi-layered distillation includes intermediate feature layers, prior embeddings, and detection head logits, ensuring CLRKDNet achieves a comparable detection accuracy to CLRNet while operating at a significantly faster speed. For further details, please see the Method section \ref{method}.



We validate the advancements of our proposed method through extensive experiments conducted on the CULane \cite{pan2018spatial} and TuSimple \cite{TuSimple} datasets, reporting state-of-the-art results on both. Additionally, comprehensive ablation studies confirm the efficacy of each component within our framework. Our principal contributions are summarized as follows:

\begin{itemize}
\item We achieved significant computational efficiency by simplifying the feature enhancement module and reducing the number of detection heads in CLRNet, resulting in an increase of up to 60\% in inference speed.
\item We introduce a novel knowledge distillation technique, wherein our streamlined student model, CLRKDNet, leverages both the intermediate feature layers, prior embeddings, and the final detection head logits of the teacher model, CLRNet, to enhance its lane detection capabilities.
\item We conducted extensive experiments across various lane detection datasets to validate the effectiveness of our proposed method, CLRKDNet, and performed comprehensive ablation studies to verify the contribution of each module to the model's performance.
\end{itemize}

\section{Related Work}

\subsection{Lane Detection}
\subsubsection{\textbf{Segmentation-based Methods}}
Segmentation-based methods treat lane detection as a pixel-wise classification task, separating lane line regions from the background. For instance, SCNN \cite{pan2018spatial} leverages a semantic segmentation framework with a message-passing mechanism to improve spatial relationships in lane detection, though its real-time application is limited by speed. Similarly, RESA \cite{zheng2021resa} enhances performance through a real-time feature aggregation module, yet remains computationally demanding due to its pixel-level processing.

\subsubsection{\textbf{Parametric Curve-based Methods}}
These methods represent lanes using curve parameters, which are then fitted to lane data. LSTR \cite{liu2021end} employs a Transformer architecture to capture both thin, long lane features and broader road features. PolyLaneNet \cite{tabelini2021polylanenet} and other works \cite{feng2022rethinking} utilize polynomial and Bezier curves, respectively, for lane detection, offering fast inference but sensitive to parameter inaccuracies.

\subsubsection{\textbf{Keypoint-based Methods}}
Keypoint-based methods detect lane lines by identifying key points and then clustering them into lane instances. PINet \cite{ko2021key} and similar approaches \cite{wang2022keypoint, qu2021focus} use advanced networks and clustering algorithms for this purpose, though requiring intensive post-processing which increases computational load.

\subsubsection{\textbf{Anchor-based Methods}}
Anchor-based methods rely on predefined line or row anchors to guide lane detection. Line-CNN \cite{li2019line} and LaneATT \cite{tabelini2021keep} use line anchors with attention mechanisms for enhanced accuracy and efficiency. Conversely, row-anchor based methods like UFLD \cite{qin2020ultra} and CondLaneNet \cite{liu2021condlanenet} offer simplicity and speed but can struggle with complex scenarios due to challenges in accurately recognizing initial lane points. CLRNet \cite{zheng2022clrnet} and its extension work \cite{honda2024clrernet} propose a cross-layer optimized lane detection network, which detects lane lines using high-level features and adjusts the lane line position using low-level features. 


\subsection{Knowledge Distillation}
Knowledge distillation involves transferring insights from a complex, often cumbersome model to a more compact and computationally efficient one, thereby enhancing the smaller model's performance and generalizability. This technique was initially introduced by Hinton \textit{et al.} in 2015 \cite{hinton2015distilling}. Over the years, it has diversified into numerous approaches, including the adoption of soft labels and tailored loss functions to refine learning processes.
In computer vision, knowledge distillation has significantly boosted the capabilities of smaller models across various tasks such as object detection, image classification, and segmentation. 
Specifically within lane detection, Hou \textit{et al.} introduced Self-Attention Distillation (SAD) \cite{hou2019learning}, which employs top-down and hierarchical attention distillation to augment representation learning and model efficacy. Notably, our distillation approach outperforms SAD, achieving an F1-score on the CULane dataset that surpasses it by nearly 10 percentage points.

\section{Methods}
\label{method}

\begin{figure*}[t!]
    \centering
    \captionsetup{font={footnotesize}}
    \includegraphics[width=.99\textwidth]{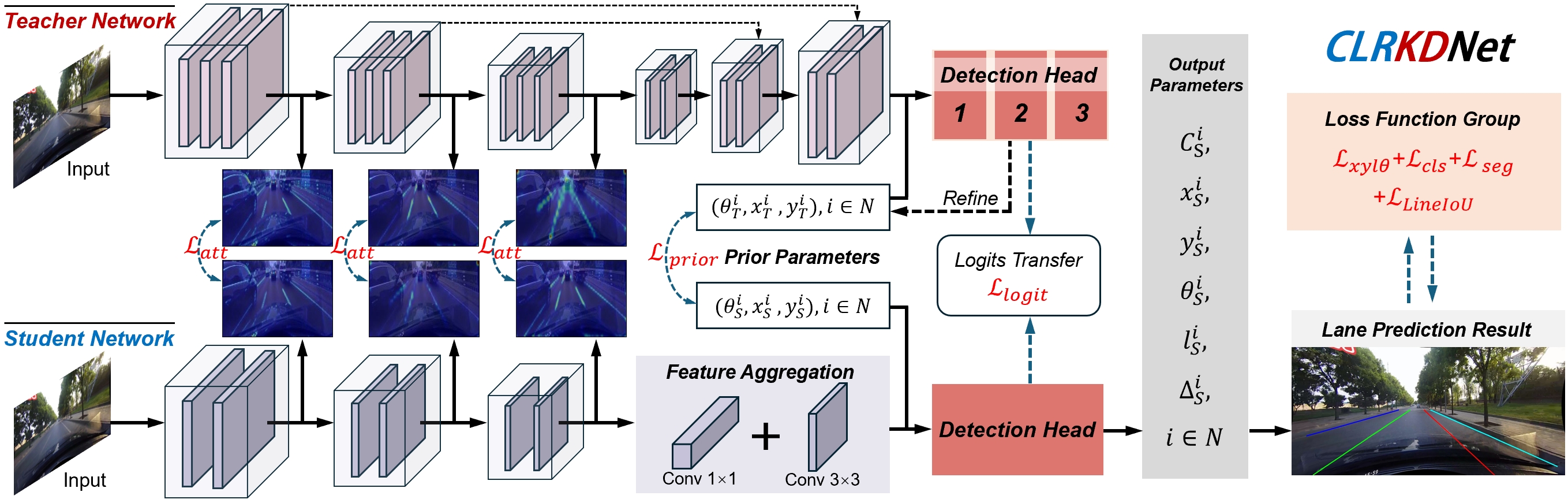}
    \vspace{-0pt}
    \caption{The upper part of the model represents the teacher's configuration, featuring a deeper backbone for feature extraction, three layers of FPN for feature fusion, and a detection head connected to each FPN layer. The detection head performs iterative prior refinement, indicated by the dashed line looping back to the prior parameters. 
    The lower part is the student network, CLRKDNet, which typically has a lighter backbone, a feature aggregation module for feature enhancement, and a single detection head for lane prediction output.  
    During the training process, three types of distillation are applied: (a) Attention Map Transfer, occurring during multi-scale feature extraction in the backbone network, transferring attention maps information from the teacher to the student model; (b) Prior Knowledge Transfer, transferring the teacher's refined priors to the student's initial priors; and (c) Logits Transfer, comparing the classification and regression outputs of both models to refine the student's performance.
 }
    \label{Model_framework}
    \vspace{-15pt}
\end{figure*}

\subsection{CLRKDNet}
\subsubsection{\textbf{Teacher Model}}

Our method utilizes the sophisticated CLRNet\cite{zheng2022clrnet} architecture as teacher model, which integrates robust backbones such as ResNet or DLA. This integration enables the backbone network to extract deep features, which are then processed by a Feature Pyramid Network (FPN) that generates multi-scale feature maps at varying resolutions, including \( \frac{1}{8}, \frac{1}{16}, \) and \( \frac{1}{32} \) of the input image size. This approach captures a comprehensive representation of both global content and local details. CLRNet initiates lane detection by configuring priors with learnable parameters \((x^i, y^i, \theta^i)\), where \((x^i, y^i)\) define the starting coordinates and \(\theta^i\) is the orientation with respect to x-axis. The symbol \( i \) represents one of the priors in a list of \( M \) priors, where \( M \) denotes the total number of priors. These priors, crucial for identifying potential lane paths, are processed through multiple convolutional and fully connected layers across various scales. As shown in Fig. \ref{Model_framework}, this multi-layer processing generates classification and regression outputs for prior adjustments. The model undergoes three refinement cycles where these adjustments recalibrate the priors, leveraging higher-resolution feature maps for improved accuracy. Cross-attention mechanisms integrate contextual information throughout this process, culminating in the precise calculation of x-coordinates across multiple horizontal rows to delineate lane paths.

To tackle the complexities of lane detection, CLRNet employs a comprehensive loss function L, combining smooth-L1 for prior refinement, focal loss for classification, and cross-entropy loss for segmentation. Additionally, a novel LineIoU loss specifically enhances the intersection-over-union metric for lane predictions, boosting the model's precision.
\begin{equation}
L = \lambda_{\text{xyl}\theta} L_{\text{xyl}\theta} + \lambda_{\text{cls}} L_{\text{cls}} + \lambda_{\text{seg}} L_{\text{seg}} + \lambda_{\text{LineIoU}} L_{\text{LineIoU}}
\label{eq:Loss_clrnet}
\end{equation}

\( L_{xyl \theta} \) denotes the smooth-L1 loss, \( L_{cls} \) is the focal loss for prior classification, \( L_{seg} \) is an auxiliary loss aiding segmentation, and \( L_{LaneIoU} \) is the specially designed IOU loss for lane line. The weighting for this loss function during training could be find in subsection \ref{subsec:training_loss}

\begin{figure}[t]
    \centering
    \captionsetup{font={footnotesize}}
    \includegraphics[width=0.45\textwidth]{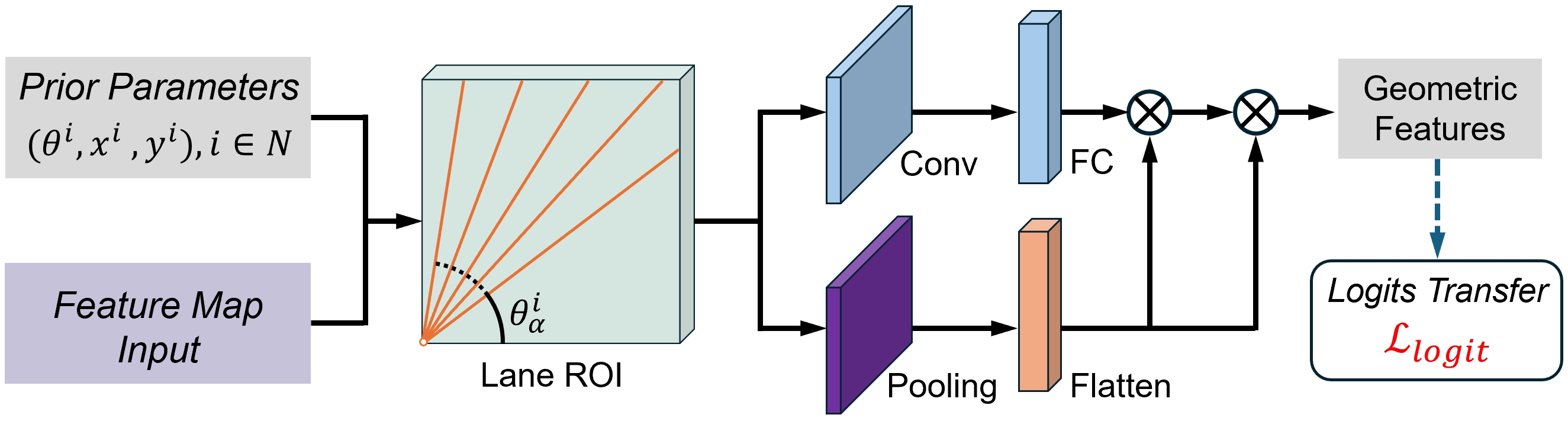}
    \vspace{-0pt}
    \caption{Illustration of detection head}
    \label{detection-head}
    \vspace{-15pt}
\end{figure}

\subsubsection{\textbf{Student Model}}
CLRKDNet is a streamlined variant of the advanced CLRNet model, designed to meet the demanding requirements of real-time autonomous driving applications while managing lane detection performance. It leverages CLRNet's advanced network design, including its backbone and detection head mechanisms, but introduces significant optimizations in its architecture to enhance efficiency.

In the feature enhancement section, where CLRNet integrates multi-scale, semantically rich features through an FPN, our CLRKDNet deploys a lean feature aggregation network to reduce computational burden. As shown in Fig. \ref{Model_framework}, this network is purpose-built to condense the channel size of the extracted features from the backbone, enhancing their representational quality without the computational heft. This innovation not only slashes the number of weight parameters but also diminishes the computational footprint, enabling a swifter feature integration process.

The detection head of our CLRKDNet is also optimized for efficiency. Unlike CLRNet’s multiple detection heads and learnable priors that require intensive computational resources for iterative refinement, our CLRKDNet adopts a single detection head with a fixed set of priors. This singular head leverages a set of static priors without engaging in costly iterative refinement processes. The structure of a single detection head is shown in Fig. \ref{detection-head}.
Experimental analysis shows that this simplification has increased the inference speed of our CLRKDNnet by up to 60\%, while the evaluation scores has only decreased slightly.



The decrease in detection performance resulting from the simplified architecture is further compensated by the knowledge distillation. Our proposed knowledge transfer method leverages distilled insights from the intermediate layers of the teacher model CLRNet, as well as priors and final output of the detection head. Through this muti-stage distillation process, our CLRKDNet model closely approaches the benchmark set by CLRNet while operating at a fraction of the computational cost.

\subsection{Attention Map Distillation}
\subsubsection{\textbf{Activation Attention Transfer}}

In our lane detection model, knowledge distillation is facilitated through an activation-based attention mechanism \cite{zagoruyko2016paying}. For each \(n\)-th layer of the convolutional neural network, we extract activation outputs, denoted as \(A_n \in \mathbb{R}^{C_n \times H_n \times W_n}\), where \(C_n\), \(H_n\), and \(W_n\) correspond to the number of channels, height, and width of the activation tensor, respectively. 

To distill knowledge from the teacher model to the student model, we generate spatial attention maps from these activation tensors. These maps serve as condensed representations that spotlight the areas within the input image deemed most critical by the model. These maps are distilled through the application of a mapping function, \( G_{\text{sum}^p}(A_n) = \sum_{j=1}^{C_n} |A_{n,j}|^p \), where each \( A_{n,j} \) denotes the \( j \)-th slice of \( A_n \) in the channel dimension, and \( p > 1 \). Drawing from the precedent set by other papers \cite{hou2019learning}, we selected this \( p = 2 \) to intensify the emphasis on the most distinct features, thereby directing the student model's focus in a manner akin to the teacher model. The process of this Attention Map Distillation is shown in Fig. \ref{Model_framework}. During the entire training phase, the attention maps of the student model are progressively adjusted to match those of the teacher model, with differences minimized using a loss function.


\subsubsection{\textbf{Attention Transfer Loss}}

The distillation of attention maps in our lane detection model is quantified through an attention transfer loss function that specifically measures the divergence between the attention maps of the student and teacher models. For each paired layer indexed by \( n \) in the set \( N \) (representing indices of every teacher-student activation layer pairs), the attention maps from the student model \( A_n^S \) and the teacher model \( A_n^T \) are first transformed into vectorized forms, denoted as \( Q_n^S \) and \( Q_n^T \), respectively. These vectorized forms are produced by applying the mapping function \( G \) in previous section to activation tensors and reshaping the resulting attention maps into vectors.
\begin{equation}
     L_{att} = \sum_{n \in N} \left\| \frac{Q_n^{S}}{\| Q_n^{S} \|_2} - \frac{Q_n^{T}}{\| Q_n^{T} \|_2} \right\|_p 
\end{equation}

where \( Q_n^{S} = \textit{vec}(G(A_n^S)) \) and \( Q_n^{T} = \textit{vec}(G(A_n^T)) \) are the vectorized forms of the \( n \)-th attention map pair between student and teacher model, respectively. The term \( \|\cdot\|_2 \) denotes the \(\ell_2\) norm, which is used to normalize each vectorized attention map, making sure that the loss computation is invariant to the scale of the attention maps and focuses purely on their patterns. The parameter \( p \) is set to 2, aligning with the use of the second degree mapping function G in attention map calculation, which has been empirically shown to facilitate effective knowledge transfer.

\subsection{Knowledge Transfer on Detection Head}
Following the attention map distillation process, we aim to further improve detection accuracy and bridge the gap between the models. To achieve this, we come up with a dual distillation procedure for the detection head, employing prior embedding distillation and logits distillation. These mechanisms ensure that CLRKDNet effectively retains the sophisticated detection capabilities of its teacher model, CLRNet, despite its streamlined architecture. The prior embedding distillation aligns the student's priors with the refined outputs of the teacher, while logits distillation measures and minimizes the discrepancies in output logits, guiding CLRKDNet's predictions to closely match those of CLRNet. 

\subsubsection{\textbf{Prior Embedding Distillation}}
Both CLRNet and CLRKDNet initialize their detection heads with embedded priors that define the geometric parameters of lane lines, including initial coordinates \( (x^i, y^i) \) and orientation \( \theta^i \) relative to the x-axis. These priors and feature map, generated by the backbone and feature fusion network, guide the ROI module for accurate gathering of nearby features for each lane pixel. This gathering process combines global content with rich semantic information, ensuring comprehensive detection capabilities. While CLRNet refines these priors across various layers of its detection head, enhancing them iteratively, CLRKDNet uses a single set of these priors directly for detection output.

The distillation of these priors is accomplished by comparing the prior embeddings between the student's initial priors and the refined priors of the teacher model. Specifically, the embedding, formatted as a tensor with \([M, 3]\) dimension where \(M\) represents the number of initial priors, are compared using an L2 norm loss function:

\begin{equation}
L_{\text{prior}} = \sum_{i=1}^M \| P_S^i - P_T^i \|_2
\end{equation}

Here, \( P_S^i \) and \( P_T^i \) represent the \( i \)-th prior vectors from the student and teacher models, respectively, each encompassing the initial coordinates and orientation \( (x^i, y^i, \theta^i) \). This L2 norm comparison measures the Euclidean distance between each pair of corresponding priors, effectively aligning CLRKDNet's static priors with the dynamically refined priors of CLRNet. This alignment ensures that the student model starts from a refinement level comparable to the output of the teacher's iterative process, effectively bridging the gap in dynamic refinement capabilities between the two models.

\begin{table*}[htbp]
\centering
\resizebox{\textwidth}{!}{%
\begin{tabular}{@{}cccccccccccccc@{}}
\toprule
Method      & Backbone & F1(\%) & FPS  & Normal & Crowd & Dazzle & Shadow & Noline & Arrow & Curve & Cross & Night \\
\midrule
SCNN \cite{pan2018spatial}       & VGG16    & 71.60  & 50   & 90.60  & 69.70  & 58.50  & 66.90   & 43.40   & 84.10  & 64.40  & 1990   & 66.10 \\
LaneATT \cite{tabelini2021keep}    & ResNet18    & 75.13  & 250  & 91.17  & 72.71  & 65.82  & 68.03   & 49.13   & 87.82  & 63.75  & 1020   & 68.58  \\
LaneATT  \cite{tabelini2021keep}   & ResNet34    & 76.68  & 170  & 92.14  & 75.03  & 64.47  & 78.15   & 49.39   & 88.38  & 67.72  & 1330   & 70.72   \\
CondLaneNet \cite{liu2021condlanenet} & ResNet18    & 78.14  & 348  & 92.87  & 75.79  & 70.72  & 80.01   & 52.39   & 89.37  & 72.40  & 1364   & 73.23   \\
CondLaneNet \cite{liu2021condlanenet} & ResNet34    & 78.74  & 237  & 93.38  & 77.14  & 71.17  & 79.93   & 51.85   & 89.89  & 73.88  & 1387   & 73.92    \\
UFLD   \cite{qin2020ultra}     & ResNet18    & 68.40  & \textbf{620}  & 87.70  & 66.00  & 58.40  & 62.80   & 40.20   & 81.00  & 57.90  & 1743   & 62.10    \\
UFLD    \cite{qin2020ultra}    & ResNet34    & 72.30  & 340  & 90.70  & 70.20  & 59.50  & 69.30   & 44.40   & 85.70  & 69.50  & 2037   & 66.79    \\
SAD \cite{hou2019learning}& ENet & 70.80 & 200 & 90.10 & 68.80 & 60.20 & 65.90 & 41.60 & 84.00 & 65.70 & 1998 & 66.00 \\
SAD \cite{hou2019learning}& ResNet101 & 71.80 & 40 & 90.70 & 70.00 & 59.90 & 67.00 & 43.50 & 84.40 & 65.70 & 2052 & 66.30 \\
\midrule
UFLDv2  \cite{qin2022ultra}    & ResNet18    & 74.70  & 330  & 91.70  & 73.00  & 64.60  & 74.70   & 47.20   & 87.60  & 68.70  & 1998   & 70.20   \\
UFLDv2   \cite{qin2022ultra}   & ResNet34    & 75.90  & 165  & 92.50  & 74.90  & 65.70  & 75.30   & 49.00   & 88.50  & 70.20  & 1864   & 70.60  \\
CLRNet  \cite{zheng2022clrnet}    & ResNet18    & 79.58  & 275  & 93.30  & 78.33  & 73.71  & 79.66   & 53.15   & 90.25  & 71.56  & 1321   & 75.11  \\
CLRNet  \cite{zheng2022clrnet} & ResNet101   & 80.13 & 46   & 93.85  & 78.78 & 72.49  & 82.33  & 54.50  & 89.79  & \textbf{75.57} & 1262  & 75.51 \\
CLRNet \cite{zheng2022clrnet}  & DLA34    & 80.47 & 185  & 93.73  & 79.59 & 75.30  & 82.51  & 54.58  & 90.62  & 74.13 & 1155  & 75.37 \\
FLAMNet \cite{ran2023flamnet} & ResNet34 & 80.15 & 93 & 93.61 & 78.35 & 73.64 & 81.31 & 53.68 & \textbf{90.67} & 73.38 & 982 & 75.50 \\
FLAMNet \cite{ran2023flamnet}& DLA34 & 80.67 & 101 & \textbf{93.94} & 79.35 & \textbf{77.02} & \textbf{82.94} & 54.62 & 90.51 & 75.23 & 1205 & 76.49 \\
\midrule

CLRKDNet (Ours)   & ResNet18    & 79.66 & 450  & 93.34 & 78.19	&74.62 &	80.46	& 52.61 & 89.98 & 69.17	& \textbf{904} & \textbf{79.66} &       \\
CLRKDNet (Ours)   & DLA34    & \textbf{80.68} & 265 & 93.86 & \textbf{79.95} & 75.63 & 81.88 & \textbf{54.85}	& 90.42 & 72.81	& 1147	& 75.82 &       \\				
\bottomrule
\end{tabular}
}
\caption{Result comparison with state-of-art methods on CULane. FPS is measured using single NVIDIA 3090 GPU}
\label{tab: CUlane_comparison}
\vspace{-15pt}
\end{table*}

\subsubsection{\textbf{Logit Distillation}}
Logit distillation focuses on the final output of the detection heads before transforming the model outputs into predicted lines. As shown in Fig.\ref{detection-head}, such logits include classification scores and geometric features, such as starting coordinates \( (x^i, y^i) \), angles \( \theta^i \), lane lengths \(l^i\), and the horizontal offset differences \(\Delta x^i\) between the predicted lanes and the lane priors. 
The logically distillation process involves comparing these logical outputs from the student's detection head against those of the teacher model and calculating a Mean Squared Error (MSE) to measure and minimize the differences between them.

This loss ensures that CLRKDNet's simplified detection head, which lacks the multiple refinement stages of CLRNet, can still produce outputs with a high degree of accuracy. The MSE, being sensitive to large discrepancies, is particularly effective in fine-tuning the student model's output to closely mimic those of the teacher, thereby compensating for the absence of iterative refinement layers. This particular loss is written as:

\begin{equation}
\begin{aligned}
L_{\text{logit}} = \frac{1}{M} \sum_{i=1}^M \Bigl( & (x^i_S - x^i_T)^2 + (y^i_S - y^i_T)^2 \\
& + (\theta^i_S - \theta^i_T)^2 + (l^i_S - l^i_T)^2 \\
& + (\Delta x^i_S - \Delta x^i_T)^2 \Bigr)
\end{aligned}
\end{equation}

Where \( M \) is the number of priors. \( x^i_S, y^i_S, \theta^i_S, l^i_S, \Delta x^i_S \) are the geometric outputs of detection head, including origin coordinates, orientation, length, and horizontal difference. \( x^i_T, y^i_T, \theta^i_T, l^i_T, \Delta x^i_T \) are the corresponding outputs from the teacher model.

\subsection{Training Details}
\label{subsec:training_loss}
\subsubsection{\textbf{Training Loss}}

\begin{itemize}
    \item Distillation Loss (\(L_{\text{dis}}\)):
    The total distillation loss is the sum of three key losses: attention map transfer loss, lane prior embeddings loss, and detection head logit loss. This comprehensive loss is formulated as follows:
    \begin{equation}
    L_{\text{dis}} = w_{\text{att}} L_{\text{att}} + w_{\text{prior}} L_{\text{prior}} + w_{\text{logit}} L_{\text{logit}}
    \label{eq:Loss_dis}
    \end{equation}
    The coefficient \(w_{\text{att}}\), \(w_{\text{prior}}\), \(w_{\text{logit}}\) adjusts the impact of the corresponding distillation loss term to optimize the transfer of knowledge from CLRNet to CLRKDNet.
    More details about distillation losses could be found in section \ref{subsubsection: Implementation_Details}

    \item Classification and Regression Loss (\(L_{\text{CR}}\)):
    For the general loss calculation, we follow the equation \ref{eq:Loss_clrnet} from CLRNet, with the loss weights set as follows: \( \lambda_{\text{xyl}\theta} = 0.2 \), \( \lambda_{\text{cls}} = 2 \), \( \lambda_{\text{seg}} = 1 \), and \( \lambda_{\text{LineIoU}} = 2 \). 

\end{itemize}

\subsubsection{\textbf{Knowledge Distillation Setups}}

For training on the CULane dataset, CLRKDNet with a ResNet18 backbone was trained on CLRNet's ResNet101 backbone model, which achieved an F1-score of 80.13\%. Meanwhile, CLRKDNet with a DLA34 backbone was trained on the retrained CLRNet's DLA34 backbone model, which attained an F1-score of 80.71\% after removing frame redundancy. For the TuSimple dataset, since ResNet18 had the best overall performance, we utilized CLRNet with a ResNet18 backbone to train CLRKDNet, maintaining the same backbone.

Following the paper by Hiroto \cite{honda2024clrernet}, redundant training data removal was conducted before training on the CULane dataset. Frames with an average pixel value difference below a threshold of 15 from the previous frame were discarded, leaving 55,698 frames (62.7\%) for training. This process mitigates overfitting and enhances CLRKDNet's baseline F1-score to 79.05\% with a ResNet18 backbone, representing an approximately 0.6\% increase from the baseline score of 78.40\% achieved by the model without redundant data removal.
With a DLA34 backbone, the F1-score of our CLRKDNet baseline model is 80.13\% after removing redundant data, which is close to the teacher model CLRNet's score of 80.47\%. After retraining CLRNet with redundant data removal, we achieve an F1-score of 80.71\%, establishing this as our new teacher model weight.

\section{EXPERIMENTS}

\subsection{Datasets}
\subsubsection{\textbf{CULane}} \cite{pan2018spatial} is an extensive dataset encompassing 88,880 training images and 34,680 test images, all formatted at a resolution of 1640x590 pixels. It covers a wide range of driving conditions, including urban and highway scenes, and categorizes test images into scenarios like night, crowded, and curved lanes.

\subsubsection{\textbf{TuSimple}} \cite{TuSimple} focuses on highway scenes and includes 3,268 training images, 358 validation images, and 2,782 testing images, each with a resolution of 1280x720 pixels. 

\subsection{Evaluation Metrics}

The efficacy of lane detection methods in our study is quantified through the metrics of Accuracy (Acc) and the F1-score.

\subsubsection{\textbf{Accuracy (Acc)}} Acc quantifies the proportion of correctly identified lane points. A prediction is deemed accurate if over 85\% of the predicted points fall within 20 pixels of the ground truth lane markers \cite{TuSimple}. The accuracy metric is computed as:
\begin{equation}
    Acc = \frac{\sum_{\textit{clip}} C_{\textit{clip}}}{\sum_{\textit{clip}} S_{\textit{clip}}}.
\end{equation}
where \(C_{\text{clip}}\) represents the number of points within the 20-pixel boundary of the ground truth, and \(S_{\text{clip}}\) is the total count of lane points in the image.

\subsubsection{\textbf{F1-score}} It is used to understand the balance between precision and recall, where predictions with an Intersection over Union (IoU) greater than a set threshold are considered as true positives (TP) \cite{pan2018spatial}. It is calculated by the formula:
\begin{equation}
    F1 = \frac{2 \times \textit{Precision} \times \textit{Recall}}{\textit{Precision} + \textit{Recall},}
\end{equation}

where the \textit{Precision} and \textit{Recall} are determined by:
\begin{equation}
    \textit{Precision} = \frac{TP}{TP+FP},
\end{equation}
\begin{equation}
    \textit{Recall} = \frac{TP}{TP+FN},
\end{equation}
with \(TP\) representing true positives, \(FP\) denoting false positives, and \(FN\) indicating false negatives.

\begin{table}[b]
\centering
\begin{tabular}{lccc}
\toprule
Method &Backbone& F1 & FPS \\
\midrule
SAD \cite{hou2019learning} &ENet& 70.80 & 200 \\
SAD \cite{hou2019learning} &ResNet101 &71.80 & 40 \\
CLRKDNet (Ours) &ResNet18& 79.66 & 450 \\
CLRKDNet (Ours) &DLA34 & 80.68 & 265 \\
\bottomrule
\end{tabular}
\caption{Performance comparison of different lane detection methods that utilize knowledge transfer}
\label{tab:distillation_compared}
\end{table}

\subsection{Implementation Details}
\label{subsubsection: Implementation_Details}
In this study, we employ knowledge distillation techniques with teacher and student models based on ResNet \cite{He_2016_CVPR} and DLA \cite{yu2018deep} architectures, pre-trained on ImageNet. Input images are resized to a consistent resolution of \(320 \times 800\) for uniform processing. Following established methods \cite{zheng2022clrnet}\cite{liu2021condlanenet}\cite{qu2021focus}, data augmentation such as random affine transformation and random horizontal flip were employed. Model optimization utilizes the AdamW \cite{loshchilov2017decoupled} optimizer with a cosine decay learning rate strategy \cite{loshchilov2016sgdr}, calibrated to initial values of \(1 \times 10^{-3}\). Training duration are set at 20 epochs for the CULane dataset and 90 epochs for the TuSimple dataset, tailored to each dataset's complexity. 

The teacher model utilizes the original CLRNet implementation, ensuring robust lane detection capabilities. All computational tasks are executed on a single RTX 3090 GPU using PyTorch.  The weighted loss of distillation loss calculation in equation \ref{eq:Loss_dis} is set as follows: \( w_{\text{att}} = 1 \), \( w_{\text{prior}} = 3 \), \( w_{\text{logit}} = 5 \). The number of initial priors \( M \) = 192 and paired activate layer \( N \)  is set to 4.


\subsection{Experimental Results}
\subsubsection{\textbf{Performance on CULane}}
We present our model's benchmark results on the CULane dataset, alongside other state-of-the-art models, in Table \ref{tab: CUlane_comparison}. Our proposed method, CLRKDNet, achieves an F1-score of 79.66\% with a ResNet18 backbone, marking a 0.08\% increase over CLRNet's score of 79.58\% with the same backbone. Additionally, the inference speed has increased from 275 FPS to 450 FPS, an impressive 60\% improvement.
With a DLA34 backbone, after removing frame redundancy and retraining the original CLRNet model, it attains an F1-score of 80.71\%. Meanwhile, CLRKDNet records an F1-score of 80.68\%, a minor decrease of 0.03\%. However, CLRKDNet's inference speed increases by approximately 80 FPS compared to the CLRNet model, from 185 FPS to 265 FPS, resulting in a 40\% increase in inference speed.

In table \ref{tab:distillation_compared}, we compare the F1-score of our model to the only other model in the field of lane detection that employs distillation. The results reveal that our model outperforms the SAD \cite{hou2019learning} model in both detection accuracy and inference speed. Despite utilizing a similarly lightweight backbone, our model nearly doubles the inference speed, and the F1-score sees an increase of almost 10\%, rising from 70.80\% to 79.66\%.

\begin{figure*}[t!]
    \centering
    \captionsetup{font={footnotesize}}
    \includegraphics[width=.99\textwidth]{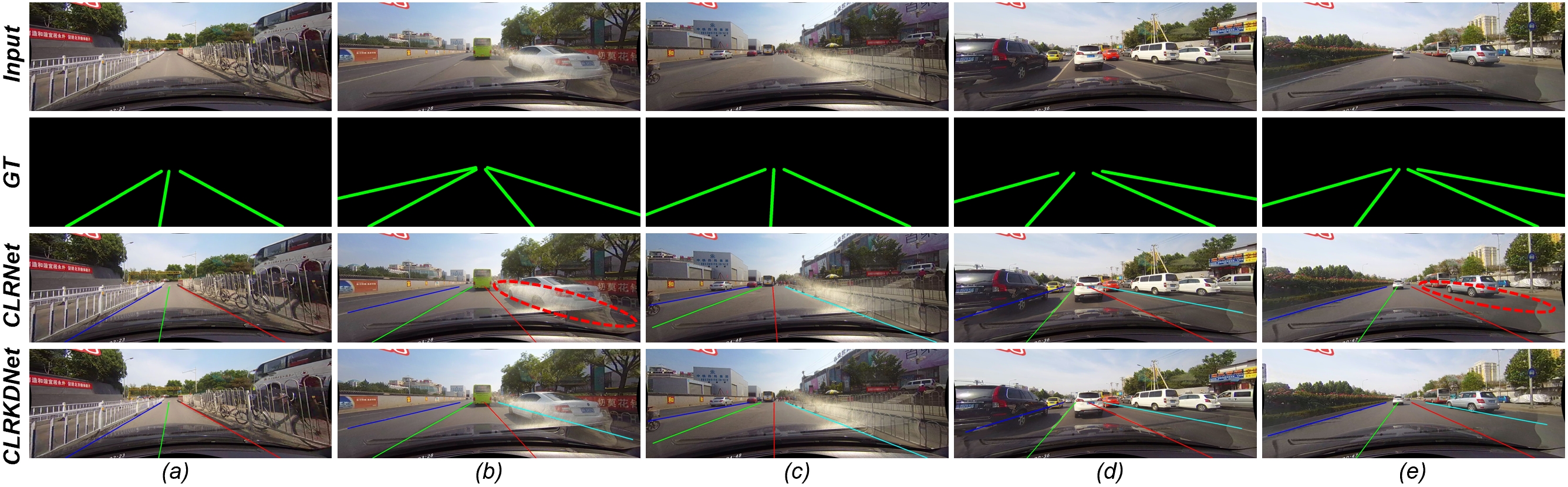}
    \vspace{-0pt}
    \caption{Selected results comparing the CLRNet teacher model with CLRKDNet student model against ground truth annotation and input images. Certain category where the student surpasses the teacher's model is shown with dotted circle representing the missing detected lane.}
    \label{img:Result_compare}
    \vspace{-15pt}
\end{figure*}

We also present a diagram in Fig. \ref{fps_vs_f1} that compares each state-of-the-art model's F1-score against its respective inference speed (FPS). The diagram highlights the balance struck by our model, CLRKDNet, between accuracy and speed. It outperforms other models in its class, achieving an additional 100-200 FPS compared to similarly performing models. The only model exhibiting higher efficiency, UFLD, achieves a significantly lower F1-score, with nearly a 10\% deficit compared to CLRKDNet. 
The visualized comparison between CLRKDNet and CLRNet is presented in Fig. \ref{img:Result_compare}. As shown in the figure, our model, CLRKDNet, achieves better results in certain categories, such as crowd circumstances, particularly in columns (b) and (d).

\subsubsection{\textbf{Performance on TuSimple}}

Table \ref{tab:TuSimple_performance} showcases the performance of our model alongside state-of-the-art approaches in TuSimple dataset. Our model, CLRKDNet, demonstrates near-parity with its teacher model, CLRNet, with only a 0.11\% drop in F1-score and a 0.07\% drop in accuracy. Despite this minimal reduction in performance metrics, our CLRKDNet achieves a runtime more than 50\% faster than its teacher model, aligning with the significant improvements observed in the CULane dataset experiments.

\begin{table}[t]
\centering
\setlength{\tabcolsep}{0pt} 
\begin{tabular*}{0.5\textwidth}{@{\extracolsep{\fill}}lccccccc@{}}
\toprule
Method & Backbone & F1(\%) & Acc(\%) & FP(\%) & FN(\%) & RT(ms) \\
\midrule
SCNN \cite{pan2018spatial} & VGG16 & 95.97 & 96.53 & 6.17 & \textbf{1.80} & 20 \\
UFLD \cite{qin2020ultra} & ResNet18 & 87.87 & 95.82 & 19.05 & 3.92 & \textbf{1.7} \\
UFLD \cite{qin2020ultra} & ResNet34 & 88.02 & 95.86 & 18.91 & 3.75 & 3.0 \\
LaneATT \cite{tabelini2021keep} & ResNet18 & 96.71 & 95.57 & 3.56 & 3.01 & 4.0 \\
LaneATT \cite{tabelini2021keep} & ResNet34 & 96.77 & 95.63 & 3.53 & 2.92 & 5.9 \\
CondLaneNet \cite{liu2021condlanenet} & ResNet18 & 97.01 & 95.48 & \textbf{2.18} & 3.80 & 2.9 \\
CondLaneNet \cite{liu2021condlanenet} & ResNet34 & 96.98 & 95.37 & 2.20 & 3.82 & 4.2 \\
UFLDv2 \cite{qin2022ultra}& ResNet18 & 96.16 & 95.65 & 3.06 & 4.61 & 3.2 \\
UFLDv2 \cite{qin2022ultra}& ResNet34 & 96.22 & 95.56 & 3.18 & 4.37 & 5.9 \\
CLRNet \cite{zheng2022clrnet} & ResNet18 & \textbf{97.89} & 96.84 & 2.28 & 1.92 & 3.7 \\
CLRNet \cite{zheng2022clrnet} & ResNet34 & 97.82 & \textbf{96.87} & 2.27 & 2.08 & 4.9 \\
\midrule
CLRKDNet (Ours) & ResNet18 & 97.78 & 96.80 & 2.43 & 1.92 & 2.4 \\
\bottomrule
\end{tabular*}
\caption{Performance comparison on TuSimple dataset}
\label{tab:TuSimple_performance}
\vspace{-15pt}
\end{table}

\subsection{Ablation Study and Analysis}

To validate the effectiveness of different components in our proposed method, we conducted an extensive experiment on the CULane dataset. We first evaluated CLRKDNet's standalone performance without distillation. This base model achieved an F1-score of 79.05\% and 80.13\% for the ResNet18 and DLA34 backbones, respectively, after accounting for frame redundancy removal.

Table \ref{tab:ablation_study_distillation} highlights the progressive improvements as various distillation losses are integrated into CLRKDNet, guiding the student model to learn from its teacher. For the ResNet18 backbone, the F1-score improves by 0.61\%, reaching 79.66\% when all distillation losses are applied. For the DLA34 backbone, the F1-score increases by 0.55\%, culminating at 80.68\%.

The differential increases between these backbones can be attributed to the depth difference during the distillation process. A deeper ResNet101 model provides more accurate distillation targets for CLRKDNet with a ResNet18 backbone, resulting in a larger performance gain compared to the DLA-based distillation, where a DLA34 teacher model is used to distill a DLA34 backboned student model.

\begin{table}[htbp]
\centering
\begin{tabular}{cccccc}
\toprule
\multicolumn{1}{c}{\textbf{Backbone}} & \multicolumn{1}{c}{$L_{\textit{att}}$} & \multicolumn{1}{c}{$L_{\textit{logit}}$} & \multicolumn{1}{c}{$L_{\textit{prior}}$} & \multicolumn{1}{c}{\textbf{F1}} \\
\midrule
ResNet18 &                       &                   &                    & 79.05 \\
ResNet18 & \checkmark             &                   &                    & 79.31 \\
ResNet18 & \checkmark             & \checkmark        &                    & 79.52 \\
ResNet18 & \checkmark             & \checkmark        & \checkmark         & 79.66 \\
\midrule
DLA34 &                        &                   &                    & 80.13 \\
DLA34 & \checkmark             &                   &                    & 80.39 \\
DLA34 & \checkmark             & \checkmark        &                    & 80.54 \\
DLA34 & \checkmark             & \checkmark        & \checkmark         & 80.68 \\
\bottomrule
\end{tabular}
\caption{Ablation study with various distillations losses.}
\label{tab:ablation_study_distillation}
\vspace{-10pt}
\end{table}

\section{CONCLUSIONS}
\label{conclusions}

This paper presents CLRKDNet, a streamlined lane detection model that effectively balances accuracy and efficiency through architectural refinements and knowledge distillation techniques. Our experiments on the CULane dataset demonstrate that CLRKDNet maintains competitive F1-score of 79.66\% and 80.68\% for ResNet18 and DLA34 backbones, respectively, while significantly improving inference speed by up to 60\%. These results highlight how the integration of distillation losses, including prior embedding distillation, attention map distillation, and logit distillation, allows CLRKDNet to replicate the sophisticated detection capabilities of its teacher model, CLRNet, despite its simplified design. This balance between accuracy and efficiency makes CLRKDNet a viable solution for real-time lane detection in autonomous driving applications.


\normalem
\printbibliography


\end{document}